\documentclass[runningheads]{llncs}
\usepackage[T1]{fontenc}
\usepackage{graphicx}
\usepackage{booktabs}
\usepackage[misc]{ifsym}
\usepackage{amsfonts}
\usepackage{amsmath}
\usepackage{algorithm}
\usepackage{algpseudocode}

\newcommand{\corr}{(\Letter)}

\begin{document}
	
	\title{LLMs can Compress LLMs: Adaptive Pruning by Agents}
	
	\titlerunning{LLMs can Compress LLMs: Adaptive Pruning by Agents}
	
	\author{Sai Varun Kodathala\inst{1} \corr \and
		Rakesh Vunnam\inst{2}}
	
	\authorrunning{S.V. Kodathala and R. Vunnam}
	
	\institute{Research and Development, Sports Vision, Inc., Minnetonka, MN 55305, USA \email{varun@sportsvision.ai}
		\and
		Research and Development, Vizworld Inc., Minnetonka, MN 55305, USA \email{rakesh@vizworld.ai}}
	
	\maketitle
	
	\begin{abstract}
		As Large Language Models (LLMs) continue to scale, post-training pruning has emerged as a promising approach to reduce computational costs while preserving performance. Existing methods such as SparseGPT and Wanda achieve high sparsity through layer-wise weight reconstruction or activation-aware magnitude pruning, but rely on uniform or hand-crafted heuristics to determine per-layer sparsity ratios. Moreover, recent work has shown that pruned LLMs suffer from severe factual knowledge degradation, with structured pruning methods experiencing near-total collapse in factual question-answering capabilities. We introduce agent-guided pruning, where a foundation model acts as an adaptive pruning agent to intelligently select which layers to prune at each iteration while preserving critical knowledge pathways. Our method constructs layer-wise sensitivity profiles by combining Wanda-inspired weight-activation metrics with gradient importance scores, normalized as z-scores for model-agnostic comparison. These statistics are processed by an LLM agent equipped with self-reflection capabilities, enabling it to learn from previous pruning outcomes and iteratively refine its strategy. A checkpoint rollback mechanism maintains model quality by reverting when perplexity degradation exceeds a threshold. We evaluate our approach on Qwen3 models (4B and 8B parameters) at approximately 45\% sparsity, demonstrating substantial improvements over structured pruning baselines: 56\% relative improvement in MMLU accuracy, 19× better factual knowledge retention on FreebaseQA, and 69\% lower perplexity degradation. Notably, our framework requires no retraining, operates in a model-agnostic manner, and exhibits effective self-correction with only 2-4 rollbacks across 21-40 iterations, demonstrating that foundation models can effectively guide the compression of other foundation models.
		
	\end{abstract}
	
	\section{Introduction}
	
	Large Language Models (LLMs) have achieved remarkable success across diverse natural language processing tasks, demonstrating emergent capabilities in reasoning, generation, and knowledge retrieval~\cite{brown2020language,touvron2023llama,touvron2023llama2}. However, their deployment remains challenging due to substantial computational and memory requirements. For instance, models with billions of parameters necessitate multiple high-memory GPUs and incur significant inference latency~\cite{frantar2023sparsegpt}, limiting their accessibility and practical applicability in resource-constrained environments.
	
	Model compression techniques, particularly neural network pruning, have emerged as promising solutions to mitigate these challenges~\cite{lecun1989optimal,han2015learning}. Post-training pruning methods offer particular appeal for LLMs, as they eliminate the need for expensive retraining on massive corpora~\cite{frantar2023sparsegpt,sun2024wanda}. Recent approaches such as SparseGPT~\cite{frantar2023sparsegpt} and Wanda~\cite{sun2024wanda} have demonstrated that LLMs can be pruned to 50\% sparsity with minimal perplexity degradation, achieving this through layer-wise weight reconstruction and activation-aware magnitude pruning, respectively. Extensions like Wanda++~\cite{yang2025wanda++} further improve upon these foundations by incorporating regional gradients at the decoder-block level.
	
	Despite these advances, existing pruning methods share a critical limitation: they rely on \textit{uniform} sparsity ratios across layers or employ hand-crafted heuristics to determine per-layer sparsity~\cite{frantar2023sparsegpt,sun2024wanda,yang2025wanda++}. This one-size-fits-all approach fails to account for the heterogeneous importance of different layers in preserving model capabilities. Moreover, recent evaluation benchmarks have revealed a severe problem with pruned LLMs: \textit{catastrophic factual knowledge degradation}. The LLM-KICK benchmark~\cite{jaiswal2024llm-kick} demonstrates that structured pruning methods experience near-total collapse in factual question-answering performance, with N:M sparsity patterns losing over 97\% accuracy on knowledge-intensive tasks even at modest sparsity ratios (25-30\%), despite negligible perplexity increases. This disconnect between perplexity and actual task performance underscores the inadequacy of current evaluation metrics and pruning strategies.
	
	In this work, we introduce \textit{agent-guided pruning}, a novel framework where a foundation model acts as an adaptive pruning oracle to intelligently determine which layers to prune at each iteration. Our approach is motivated by two key insights: First, LLMs possess sophisticated reasoning capabilities that can be leveraged for optimization tasks beyond their traditional use in text generation~\cite{wei2022chain,yao2023tree}. Second, the heterogeneous sensitivity of layers to pruning necessitates adaptive, context-aware decision-making rather than fixed heuristics~\cite{ma2023llm-pruner}.
	
	Our method constructs comprehensive layer-wise sensitivity profiles by combining Wanda-inspired weight-activation metrics~\cite{sun2024wanda} with gradient importance scores, normalized as z-scores for model-agnostic comparison. These statistics, along with iterative feedback from previous pruning outcomes, are provided to an LLM agent equipped with self-reflection capabilities. The agent learns from past decisions, reasoning about which layers preserve critical knowledge pathways and which can safely be pruned. A checkpoint rollback mechanism serves as a safety net, reverting the model when perplexity degradation exceeds a specified threshold and penalizing the agent through the feedback loop.
	
	We evaluate our approach on Qwen3 models (4B and 8B parameters) at approximately 45\% sparsity, comparing against structured pruning baselines. Our results demonstrate substantial improvements: 56\% relative improvement in MMLU accuracy, 19× better factual knowledge retention on FreebaseQA, and 69\% lower perplexity degradation compared to the best performing 4:8 structured baseline. Notably, our framework requires no retraining, operates in a model-agnostic manner, and exhibits effective self-correction with only 9.5-10\% rollback rates across 21-40 pruning iterations. These results establish that foundation models can effectively guide the compression of other foundation models, opening a new paradigm for automated neural architecture optimization.
	
	\subsection{Main Contributions}
	
	The main contributions of this work are as follows:
	
	\begin{itemize}
		\item We introduce the first framework to use an LLM as an adaptive pruning agent, enabling intelligent, iterative layer selection for neural network compression without manual heuristic design.
		
		\item We develop a self-reflection mechanism that enables the pruning agent to learn from previous decisions, progressively refining its strategy through iterative feedback.
		
		\item We propose a model-agnostic statistical profiling approach using z-score normalization of Wanda metrics and gradient importance, enabling comparison across heterogeneous layer types.
		
		\item We demonstrate a checkpoint rollback mechanism that maintains model quality by reverting unsuccessful pruning attempts, with remarkably low rollback rates (9.5-10\%) indicating effective agent learning.
		
		\item We achieve substantial empirical improvements over structured pruning baselines, particularly in preserving factual knowledge (19× improvement) and maintaining general task performance (56\% MMLU improvement), directly addressing the critical knowledge degradation problem revealed by recent benchmarks~\cite{jaiswal2024llm-kick}.
	\end{itemize}

	\section{Related Work}
	
	\subsection{Post-Training Pruning for Large Language Models}
	
	The computational demands of billion-parameter LLMs have motivated extensive research in post-training compression methods that avoid expensive retraining. \textbf{SparseGPT}~\cite{frantar2023sparsegpt} pioneered one-shot pruning for massive language models by framing the problem as layer-wise sparse regression, using second-order information to reconstruct weights after pruning. The method achieves 50-60\% unstructured sparsity on models like OPT-175B and BLOOM-176B in under 4.5 hours, with minimal perplexity increase. However, SparseGPT's reliance on Hessian approximations incurs significant computational overhead and memory requirements.
	
	\textbf{Wanda}~\cite{sun2024wanda} introduced a simpler alternative, pruning weights by the product of their magnitudes and input activation norms on a per-output basis. Motivated by emergent large-magnitude features in LLMs~\cite{dettmers2022llm}, Wanda eliminates the need for weight updates or second-order information, achieving competitive performance with SparseGPT while requiring no retraining. The method demonstrates that effective sparse subnetworks exist exactly within pretrained LLMs without modification, and shows that pruning becomes more effective as model size increases, with 50\% sparse LLaMA-65B matching the zero-shot performance of its dense counterpart.
	
	Recent work has sought to enhance these methods through gradient information. \textbf{Wanda++}~\cite{yang2025wanda++} introduces regional gradients computed at the decoder-block level, combined with regional optimization to minimize pruning-induced output discrepancies. This achieves up to 32\% perplexity improvement over Wanda on language modeling tasks while maintaining efficiency (pruning a 7B model in under 10 minutes on a single H100 GPU).
	
	\subsection{Structured Pruning Approaches}
	
	While unstructured pruning achieves high sparsity, it often fails to deliver practical speedups on existing hardware~\cite{xiao2023smoothquant}. \textbf{Structured pruning} methods remove entire architectural components (attention heads, neurons, layers), enabling direct acceleration without specialized hardware~\cite{ma2023llm-pruner}.
	
	\textbf{LLM-Pruner}~\cite{ma2023llm-pruner} performs task-agnostic structural pruning by identifying coupled structures through dependency detection and estimating their importance using first-order gradients and Hessian approximations. The method enables efficient recovery through LoRA tuning~\cite{hu2021lora} in merely 3 hours with 50K samples, achieving 94.97\% performance retention at 20\% parameter reduction. However, the approach still relies on simplified pairwise neuron dependencies and requires manual specification of pruning ratios.
	
	N:M semi-structured sparsity patterns (e.g., 2:4, 4:8) offer a middle ground, achieving hardware acceleration through regular sparse patterns while maintaining relatively high density~\cite{frantar2023sparsegpt,sun2024wanda}. Both SparseGPT and Wanda support these patterns, generalizing their pruning criteria to enforce zero constraints in blocks of M consecutive weights.
	
	\subsection{Evaluation of Compressed LLMs}
	
	A critical issue highlighted by recent work is the inadequacy of perplexity as an evaluation metric for compressed LLMs. \textbf{LLM-KICK}~\cite{jaiswal2024llm-kick} introduced a comprehensive benchmark revealing that all pruning methods suffer significant performance degradation on knowledge-intensive tasks, even at trivial sparsity ratios (25-30\%), despite negligible perplexity increases. The benchmark demonstrates that structured N:M pruning methods experience catastrophic failure on factual question-answering (e.g., FreebaseQA), losing over 97\% of accuracy, while maintaining reasonable perplexity scores. This finding challenges the prevailing assumption that perplexity adequately captures model capability degradation and motivates our focus on preserving factual knowledge through adaptive layer selection.
	
	Beyond factual knowledge, recent studies have identified a dichotomy in pruning effects: while parametric knowledge degrades predictably, instruction-following capabilities can paradoxically improve with pruning~\cite{martra2024fragile}, suggesting complex trade-offs in how compression affects different cognitive functions.
	
	\subsection{Learning-Based Compression and Meta-Optimization}
	
	Our work is inspired by recent advances in using learned policies for neural architecture decisions. AutoML approaches have demonstrated that learned strategies can outperform hand-crafted heuristics for architecture search~\cite{zoph2016neural,real2019regularized}. More recently, LLMs have been applied to optimization tasks beyond text generation, including code generation~\cite{chen2021codex}, theorem proving~\cite{trinh2024solving}, and mathematical reasoning~\cite{wei2022chain}. 
	
	The concept of using foundation models as agents for optimization tasks has gained traction through work on tool use~\cite{schick2023toolformer}, planning~\cite{yao2023tree}, and self-reflection~\cite{shinn2023reflexion}. These works demonstrate that LLMs can engage in complex reasoning about abstract spaces and learn from feedback. However, to our knowledge, no prior work has applied LLM agents to the neural network pruning problem. Our approach extends this paradigm by showing that foundation models can effectively guide the compression of other foundation models, opening new possibilities for automated model optimization.
	
	Unlike prior pruning methods that rely on fixed importance metrics or uniform compression strategies, our agent-guided framework adapts its strategy based on observed outcomes, embodying a form of meta-learning where the pruning policy itself is refined through experience. This is particularly valuable for the heterogeneous layer sensitivities observed in modern LLMs, where different components play vastly different roles in preserving capabilities.
	
	\section{Method}
	
	\subsection{Overview}
	
	Our agent-guided pruning framework consists of four key components: (1) layer-wise sensitivity profiling using activation and gradient statistics, (2) an LLM agent that iteratively selects layers to prune based on normalized metrics, (3) a self-reflection mechanism that enables the agent to learn from previous decisions, and (4) a checkpoint rollback system that maintains model quality. Unlike prior work that applies uniform or hand-crafted sparsity ratios, our method adaptively determines per-layer pruning amounts through learned decision-making.
	
	\subsection{Layer Sensitivity Profiling}
	
	For each linear layer $\ell$ in the model, we construct a comprehensive sensitivity profile combining multiple statistical measures. Following Wanda~\cite{sun2024wanda}, we compute the weight-activation metric:
	
	\begin{equation}
		\mathbf{S}_\ell = |\mathbf{W}_\ell| \odot \|\mathbf{X}_\ell\|_2
	\end{equation}
	
	where $\mathbf{W}_\ell \in \mathbb{R}^{d_{out} \times d_{in}}$ is the weight matrix, $\mathbf{X}_\ell \in \mathbb{R}^{N \times d_{in}}$ are the input activations collected over $N$ calibration samples, $|\cdot|$ denotes element-wise absolute value, $\odot$ is the Hadamard product, and $\|\cdot\|_2$ computes the $\ell_2$-norm across samples. The sensitivity score for layer $\ell$ is defined as the $k$-th percentile of active (non-zero) weights in $\mathbf{S}_\ell$, where $k=10$ in our experiments.
	
	To capture gradient information inspired by Wanda++~\cite{yang2025wanda++}, we compute gradient importance as:
	
	\begin{equation}
		\mathbf{G}_\ell = \frac{1}{M} \sum_{i=1}^{M} |\nabla_{\mathbf{W}_\ell} \mathcal{L}_i|
	\end{equation}
	
	where $\mathcal{L}_i$ is the language modeling loss on calibration sample $i$ and $M$ is the number of gradient samples. We collect gradients every third iteration to balance computational cost with information gain.
	
	For model-agnostic comparison across heterogeneous layers, we normalize both metrics using z-score standardization:
	
	\begin{equation}
		z_\ell^{(s)} = \frac{s_\ell - \mu_s}{\sigma_s + \epsilon}, \quad z_\ell^{(g)} = \frac{g_\ell - \mu_g}{\sigma_g + \epsilon}
	\end{equation}
	
	where $s_\ell$ and $g_\ell$ are the raw sensitivity and gradient scores for layer $\ell$, $\mu$ and $\sigma$ denote mean and standard deviation across all layers, and $\epsilon = 10^{-9}$ prevents division by zero. Negative z-scores indicate below-average sensitivity (safer to prune), while positive z-scores indicate above-average sensitivity (riskier to prune).
	
	The complete profile for layer $\ell$ is: $\mathcal{P}_\ell = \{z_\ell^{(s)}, z_\ell^{(g)}, \rho_\ell\}$, where $\rho_\ell$ is the current sparsity ratio of layer $\ell$.
	
	\subsection{LLM Agent Design}
	
	At each iteration $t$, we query a foundation model \textit{(gemini-3-flash-preview)} to select layers for pruning. The agent receives:
	
	\begin{itemize}
		\item Current global sparsity $\rho_t$ and target sparsity $\rho^*$
		\item Layer profiles $\{\mathcal{P}_\ell\}_{\ell=1}^L$ sorted by sensitivity z-score
		\item Current and baseline perplexity: $\text{PPL}_t$, $\text{PPL}_0$
		\item Feedback summary from iteration $t-1$ (if $t > 1$)
	\end{itemize}
	
	The agent is instructed to reason about which layers are safe to prune based on the statistical profiles, considering both the gap remaining to target sparsity and the model's current health (perplexity degradation). The agent outputs structured JSON containing:
	
	\begin{itemize}
		\item \texttt{reasoning}: Natural language explanation of the pruning strategy
		\item \texttt{stop\_pruning}: Boolean indicating whether to terminate
		\item \texttt{layer\_decisions}: List of (layer name, additional sparsity) pairs
	\end{itemize}
	
	where additional sparsity $\delta_\ell \in [0.01, 0.15]$ specifies how much additional sparsity to induce in layer $\ell$. We use structured output with JSON schema to ensure reliable parsing.
	
	\subsection{Self-Reflection Mechanism}
	
	To enable the agent to learn from its decisions, we implement an iterative feedback loop. After pruning at iteration $t$, we compute:
	
	\begin{itemize}
		\item Sparsity gain: $\Delta\rho_t = \rho_{t+1} - \rho_t$
		\item Perplexity change: $\Delta_{\text{PPL}} = \frac{\text{PPL}_{t+1} - \text{PPL}_t}{\text{PPL}_t} \times 100\%$
	\end{itemize}
	
	At iteration $t+1$, the agent receives a feedback summary containing:
	\begin{itemize}
		\item Its previous reasoning and layer selections
		\item The observed sparsity gain and perplexity change
		\item A qualitative assessment (e.g., "Excellent - High sparsity gain with minimal PPL impact")
	\end{itemize}
	
	This feedback enables the agent to recognize effective patterns (e.g., prioritizing layers with highly negative z-scores) and adjust its strategy (e.g., becoming more conservative as perplexity degrades). The system prompt explicitly instructs the agent to analyze past decisions and refine its approach accordingly.
		\begin{algorithm}[t]
			\caption{LLM-Guided Pruning with Self-Reflection}
			\label{alg:llm-pruning}
			\begin{algorithmic}[1]
				\Require Model $\mathcal{M}$, calibration data $\mathcal{D}$, target sparsity $s_{\text{target}}$
				\Require LLM $\mathcal{L}$, rollback threshold $\tau_{\text{rollback}}$
				\State Compute baseline perplexity $\text{PPL}_0$ on $\mathcal{D}$
				\State Initialize checkpoint $\mathcal{C} \gets \emptyset$, iteration memory $\mathcal{H} \gets \emptyset$
				\For{$t = 1, \ldots, T$}
				\State Save checkpoint $\mathcal{C} \gets \{\mathbf{W}^{(l)}\}_{l=1}^{L}$, $\text{PPL}_{\mathcal{C}} \gets \text{PPL}_{t-1}$
				\State Collect activations $\{\mathbf{A}^{(l)}\}$ and gradients $\{\mathbf{G}^{(l)}\}$ from $\mathcal{D}$ \hfill \textit{// Wanda++ metrics}
				\For{layer $l = 1, \ldots, L$}
				\State Compute sensitivity: $\mathbf{S}^{(l)} = |\mathbf{W}^{(l)}| \odot \mathbf{A}^{(l)}$
				\State Compute z-scores: $z_{\text{sens}}^{(l)}, z_{\text{grad}}^{(l)}$ from $\mathbf{S}^{(l)}, \mathbf{G}^{(l)}$
				\EndFor
				\State Query LLM: $\pi_t \gets \mathcal{L}(\{z_{\text{sens}}^{(l)}, z_{\text{grad}}^{(l)}, s_t^{(l)}\}, \mathcal{H}_{t-1})$ \hfill \textit{// Layer stats + feedback}
				\If{$\pi_t.\text{stop} = \text{true}$}
				\State \textbf{break}
				\EndIf
				\For{$(l, \Delta s) \in \pi_t.\text{decisions}$}
				\State Prune layer $l$ by sparsity $\Delta s$ using $\mathbf{S}^{(l)}$ (Wanda)
				\EndFor
				\State Compute new perplexity $\text{PPL}_t$ and sparsity $s_t$
				\If{$\text{PPL}_t / \text{PPL}_{t-1} > \tau_{\text{rollback}}$}
				\State Restore checkpoint: $\{\mathbf{W}^{(l)}\} \gets \mathcal{C}$, $\text{PPL}_t \gets \text{PPL}_{\mathcal{C}}$ \hfill \textit{// Rollback}
				\EndIf
				\State Store feedback: $\mathcal{H}_t \gets \{\pi_t, s_{t-1}, s_t, \text{PPL}_{t-1}, \text{PPL}_t\}$ \hfill \textit{// Self-reflection}
				\If{$s_t \geq s_{\text{target}}$}
				\State \textbf{break}
				\EndIf
				\EndFor
				\State \Return Pruned model $\mathcal{M}$
			\end{algorithmic}
		\end{algorithm}
	
	\subsection{Checkpoint Rollback Mechanism}
	
	To prevent catastrophic degradation, we implement a safety mechanism that monitors perplexity changes. Before each pruning operation, we save the current model state. After pruning, if:
	
	\begin{equation}
		\frac{\text{PPL}_{t+1} - \text{PPL}_t}{\text{PPL}_t} > \tau
	\end{equation}
	
	where $\tau = 0.15$ (15\% threshold), we rollback to the previous checkpoint, discard the current iteration's decisions, and provide negative feedback to the agent. This rollback is communicated through the self-reflection loop with the assessment "Poor - Excessive PPL degradation, consider more conservative approach."
	
	\subsection{Complete Algorithm}
	
	Algorithm~\ref{alg:llm-pruning} presents the complete agent-guided pruning procedure.

	\section{Experiments}
	
	\subsection{Experimental Setup}
	
	\textbf{Models.} We evaluate our method on two Qwen3 models: Qwen3-4B and Qwen3-8B~\cite{qwen2024}. These models represent different scales within the same architectural family, enabling us to assess generalization across model sizes.
	
	\textbf{Baselines.} We compare against two structured pruning methods that achieve similar sparsity levels:
	\begin{itemize}
		\item \textbf{2:4 Structured Pruning}: Prunes 2 out of every 4 consecutive weights, achieving $\sim$42-45\% sparsity. This pattern is hardware-efficient on NVIDIA GPUs with Ampere architecture and beyond.
		\item \textbf{4:8 Structured Pruning}: Prunes 4 out of every 8 consecutive weights, also achieving $\sim$42-45\% sparsity with better density than 2:4.
	\end{itemize}
		\begin{table}[t]
		\caption{MMLU performance by category for Qwen3-8B. Our agent-guided method maintains substantially better performance than structured baselines across all knowledge domains, with Social Sciences showing the strongest retention at 79.2\% of baseline performance.}\label{tab:mmlu_categories}
		\centering
		\small
		\begin{tabular}{lcccc}
			\toprule
			\textbf{Category} & \textbf{BASE} & \textbf{2:4} & \textbf{4:8} & \textbf{Ours} \\
			& (\%) & (\%) & (\%) & (\%) \\
			\midrule
			STEM & 74.20 & 31.51 & 34.83 & \textbf{52.00} \\
			Humanities & 75.85 & 28.95 & 34.37 & \textbf{54.42} \\
			Social Sciences & 82.04 & 32.87 & 38.50 & \textbf{64.97} \\
			Other & 77.60 & 31.59 & 38.09 & \textbf{58.75} \\
			\midrule
			\textbf{Overall} & \textbf{77.38} & \textbf{31.35} & \textbf{36.29} & \textbf{56.67} \\
			\bottomrule
		\end{tabular}
	\end{table}
	
	Both baselines use magnitude-based pruning as implemented in standard pruning libraries. We do not include unstructured magnitude pruning or SparseGPT as baselines because recent work~\cite{jaiswal2024llm-kick} has shown that while these methods achieve good perplexity scores, they suffer from even worse factual knowledge degradation than structured methods.
	
	\textbf{Evaluation Protocol.} We follow the LLM-KICK benchmark~\cite{jaiswal2024llm-kick} evaluation protocol:
	\begin{itemize}
		\item \textbf{MMLU}~\cite{hendrycks2021measuring}: 5-shot evaluation on 57 subjects covering STEM, humanities, social sciences, and other domains. We report overall accuracy and per-category breakdowns.
		\item \textbf{FreebaseQA}~\cite{jiang2019freebaseqa}: Factual question-answering on 20,358 questions. This metric directly measures factual knowledge retention.
		\item \textbf{WikiText-2 Perplexity}~\cite{merity2016pointer}: Language modeling performance on the full WikiText-2 test set.
	\end{itemize}
	
	All evaluations use the full datasets without truncation to ensure comprehensive assessment.
	
	\textbf{Implementation Details.} We use 128 calibration samples of sequence length 2048 from the C4 dataset~\cite{raffel2020exploring}. The LLM agent is \textit{gemini-3-flash-preview} with temperature 0.5 to balance creativity and consistency. We set target sparsity to 50\%, though the algorithm may stop earlier if the agent determines further pruning would be too harmful. Activation collection uses 16 samples, gradient collection uses 8 samples, and perplexity evaluation uses 32 samples. The rollback threshold is $\tau = 0.15$ (15\% perplexity increase). All experiments run on a single NVIDIA A100 80GB GPU.
	
	\subsection{Main Results}
	
	Tables~\ref{tab:results_8b} and~\ref{tab:results_4b} present the comprehensive results for Qwen3-8B and Qwen3-4B respectively.
	
	\begin{table}[t]
		\caption{Comprehensive evaluation on Qwen3-8B at $\sim$43\% sparsity. Our agent-guided method substantially outperforms structured pruning baselines across all metrics, particularly in factual knowledge retention (19× improvement over 4:8).}\label{tab:results_8b}
		\centering
		\small
		\begin{tabular}{lcccc}
			\toprule
			\textbf{Method} & \textbf{Sparsity} & \textbf{MMLU} & \textbf{FreebaseQA} & \textbf{Perplexity} \\
			& (\%) & (\%) & (\%) & (WikiText-2) \\
			\midrule
			BASE (Dense) & 0.00 & 77.38 & 50.56 & 9.72 \\
			\midrule
			2:4 Structured & 42.40 & 31.35 & 0.22 & 103.01 \\
			4:8 Structured & 42.40 & 36.29 & 1.33 & 60.67 \\
			\textbf{Ours (Agent-Guided)} & \textbf{43.02} & \textbf{56.67} & \textbf{25.16} & \textbf{19.06} \\
			\midrule
			\multicolumn{5}{l}{\textit{Relative Improvement over Best Baseline (4:8):}} \\
			\textbf{Ours vs 4:8} & +0.62 & \textbf{+56.2\%} & \textbf{+1791\%} & \textbf{-68.6\%} \\
			\bottomrule
		\end{tabular}
	\end{table}
	
	\begin{table}[t]
		\caption{Comprehensive evaluation on Qwen3-4B at $\sim$45\% sparsity. Our method demonstrates consistent improvements over structured baselines, with particular strength in preserving general task performance (MMLU) and reducing perplexity degradation.}\label{tab:results_4b}
		\centering
		\small
		\begin{tabular}{lcccc}
			\toprule
			\textbf{Method} & \textbf{Sparsity} & \textbf{MMLU} & \textbf{FreebaseQA} & \textbf{Perplexity} \\
			& (\%) & (\%) & (\%) & (WikiText-2) \\
			\midrule
			BASE (Dense) & 0.00 & 71.29 & 32.43 & 13.64 \\
			\midrule
			2:4 Structured & 45.16 & 26.04 & 0.20 & 319.75 \\
			4:8 Structured & 45.16 & 29.24 & 0.51 & 81.28 \\
			\textbf{Ours (Agent-Guided)} & \textbf{45.58} & \textbf{44.43} & \textbf{2.08} & \textbf{39.40} \\
			\midrule
			\multicolumn{5}{l}{\textit{Relative Improvement over Best Baseline (4:8):}} \\
			\textbf{Ours vs 4:8} & +0.42 & \textbf{+51.9\%} & \textbf{+308\%} & \textbf{-51.5\%} \\
			\bottomrule
		\end{tabular}
	\end{table}
	
	For Qwen3-8B at 43\% sparsity, our method achieves 56.67\% MMLU accuracy, representing a 56.2\% relative improvement over the 4:8 structured baseline. More dramatically, we retain 25.16\% FreebaseQA accuracy compared to just 1.33\% for 4:8—a 19× improvement that directly addresses the catastrophic factual knowledge collapse identified by LLM-KICK~\cite{jaiswal2024llm-kick}. Perplexity increases by only 96\% compared to 524\% for 4:8, demonstrating substantially better language modeling preservation.
	
	For Qwen3-4B at 45.58\% sparsity, we observe consistent patterns: 51.9\% relative MMLU improvement (44.43\% vs 29.24\%), 4.1× better FreebaseQA retention (2.08\% vs 0.51\%), and 51.5\% lower perplexity degradation. These results demonstrate that agent-guided pruning generalizes effectively across model scales.
	
	Table~\ref{tab:mmlu_categories} presents the detailed MMLU breakdown by category for Qwen3-8B. Our method preserves performance across all categories, with particularly strong results in Social Sciences (79.2\% of baseline) and Humanities (71.7\% of baseline).
		\begin{figure}[t]
		\centering
		\includegraphics[width=\textwidth]{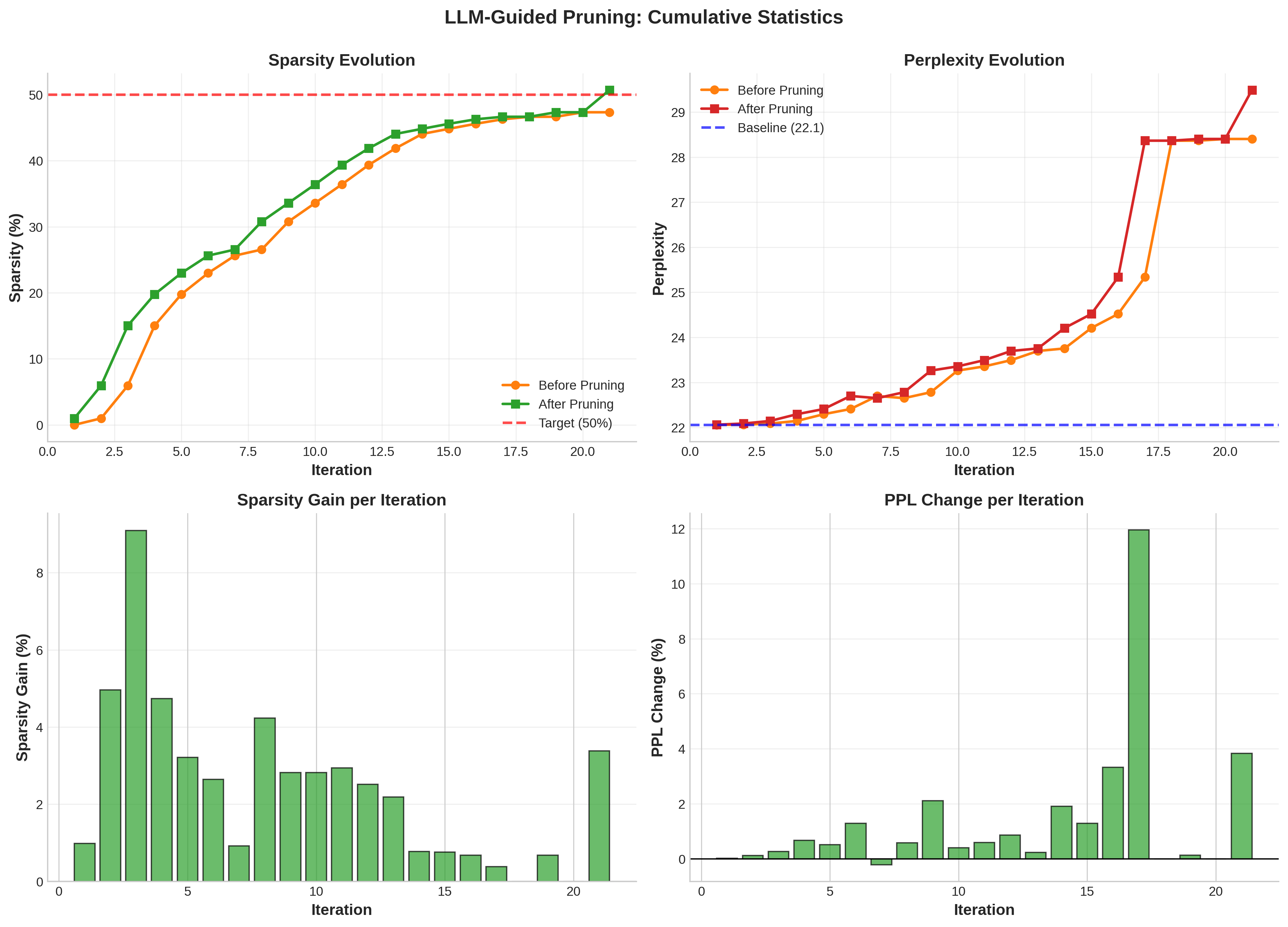}
		\caption{Cumulative statistics for agent-guided pruning on Qwen3-8B across 21 iterations. Top left: Sparsity evolution showing gradual progression to 50\% target. Top right: Perplexity evolution showing controlled degradation with 2 rollback events. Bottom left: Per-iteration sparsity gains, with most iterations achieving 1-3\% progress. Bottom right: Per-iteration perplexity changes, showing the agent learns to keep PPL increases below 2\% in most iterations.}
		\label{fig:cumulative_8b}
	\end{figure}
	
	\begin{figure}[t]
		\centering
		\includegraphics[width=\textwidth]{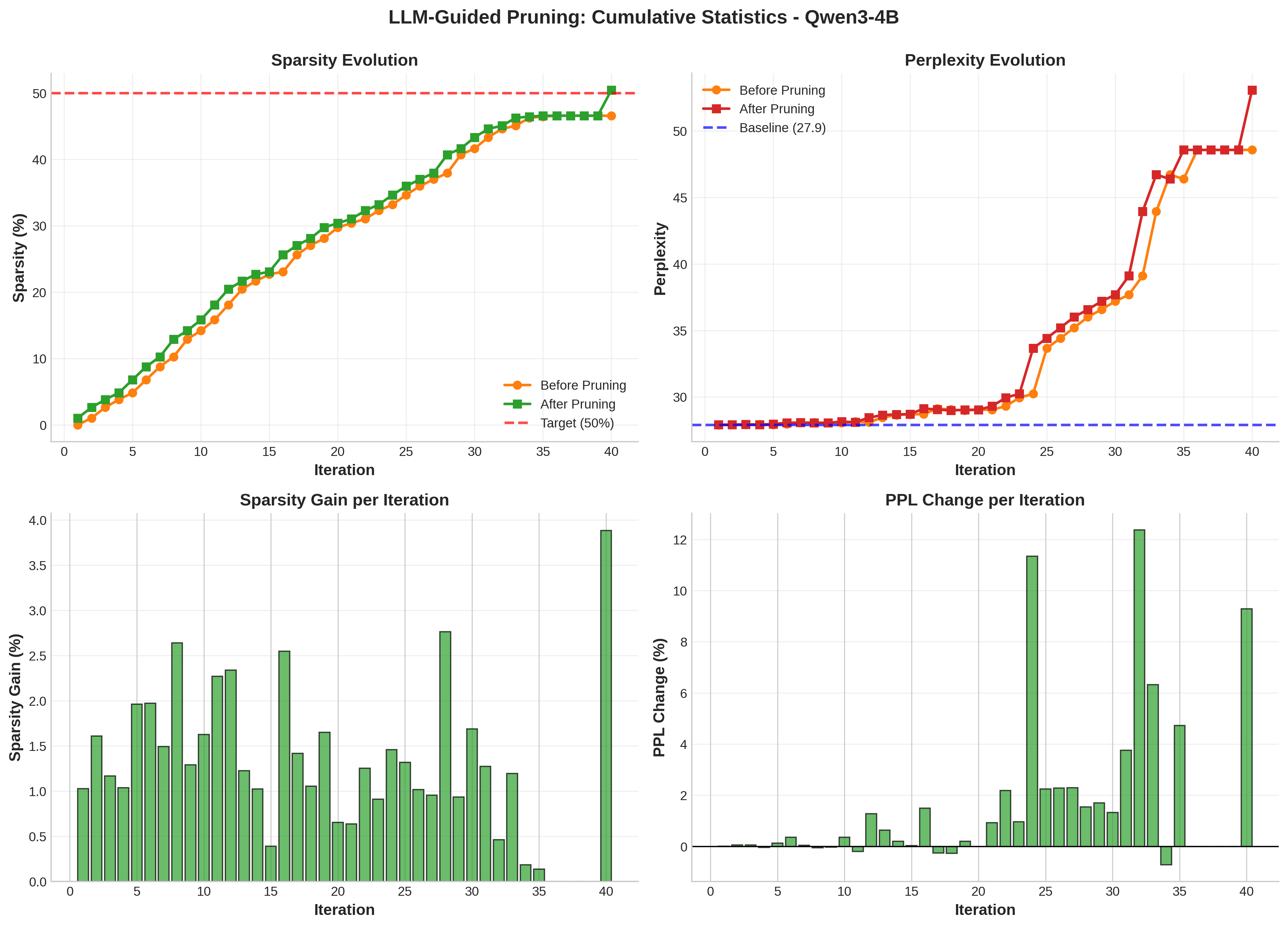}
		\caption{Cumulative statistics for agent-guided pruning on Qwen3-4B across 40 iterations. The agent achieves larger sparsity gains (3-9\%) in early iterations when the model is robust, then becomes more conservative as perplexity rises. Four rollback events (iterations 10, 15, 25, 32) are followed by visible strategy adjustments, demonstrating effective learning from negative feedback.}
		\label{fig:cumulative_4b}
	\end{figure}
	\begin{figure}[t]
		\centering
		\includegraphics[width=\textwidth]{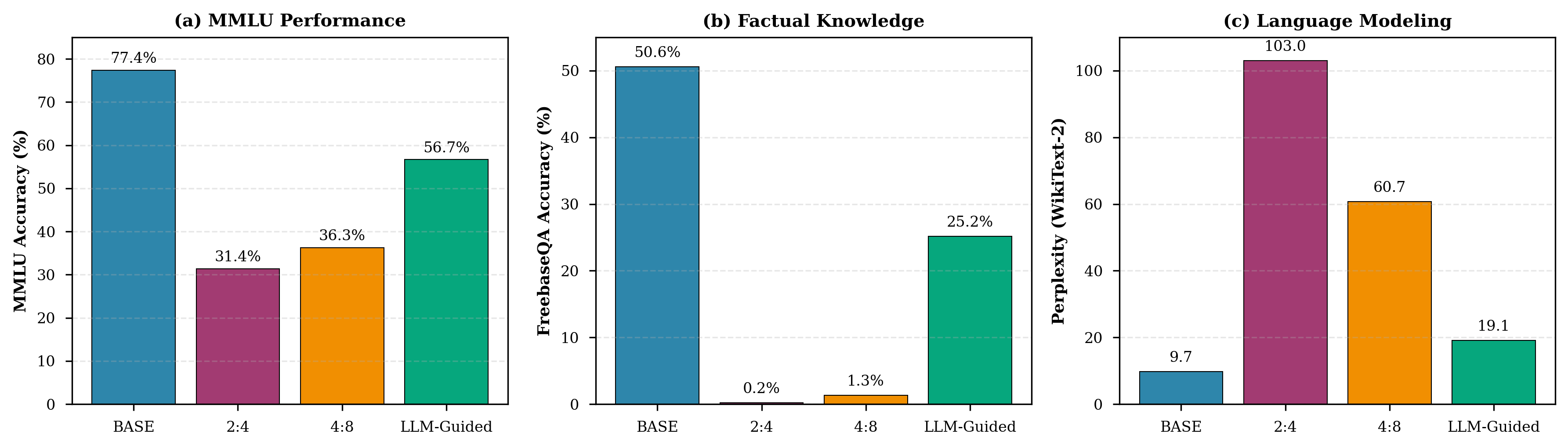}
		\caption{Performance comparison on Qwen3-8B at $\sim$43\% sparsity across three evaluation metrics. Our agent-guided method substantially outperforms structured pruning baselines, particularly in preserving factual knowledge (FreebaseQA) where structured methods experience catastrophic degradation.}
		\label{fig:comparison_8b}
	\end{figure}
	
	\begin{figure}[t]
		\centering
		\includegraphics[width=\textwidth]{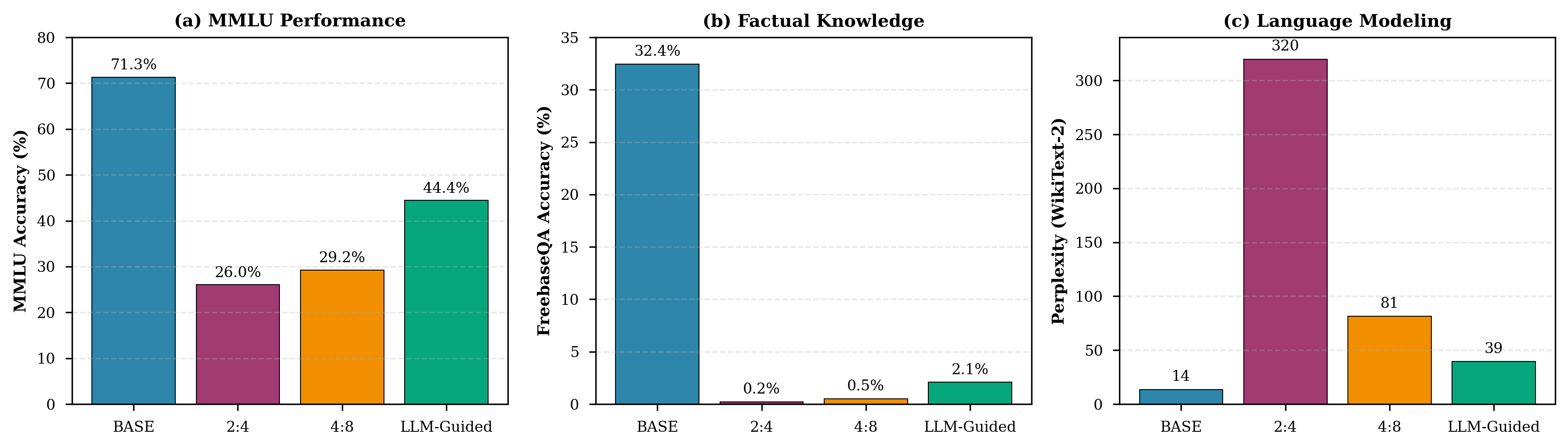}
		\caption{Performance comparison on Qwen3-4B at $\sim$45\% sparsity. The pattern of improvements is consistent with the 8B model, demonstrating that agent-guided pruning generalizes effectively across model scales.}
		\label{fig:comparison_4b}
	\end{figure}
	
	\begin{figure}[t]
		\centering
		\includegraphics[width=0.85\textwidth]{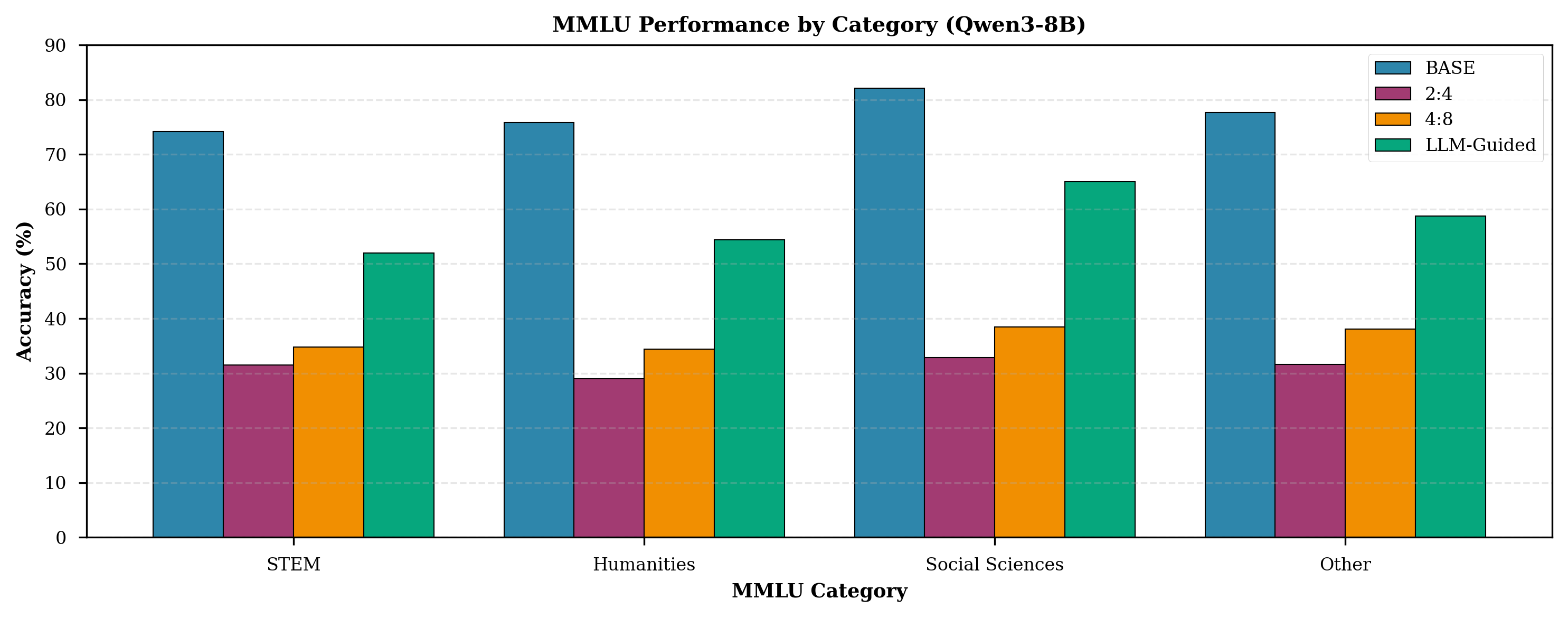}
		\caption{MMLU performance by category for Qwen3-8B. Our method maintains substantially better performance than structured baselines across all knowledge domains, with Social Sciences showing the strongest retention.}
		\label{fig:mmlu_categories}
	\end{figure}
	
	\begin{figure}[t]
		\centering
		\includegraphics[width=\textwidth]{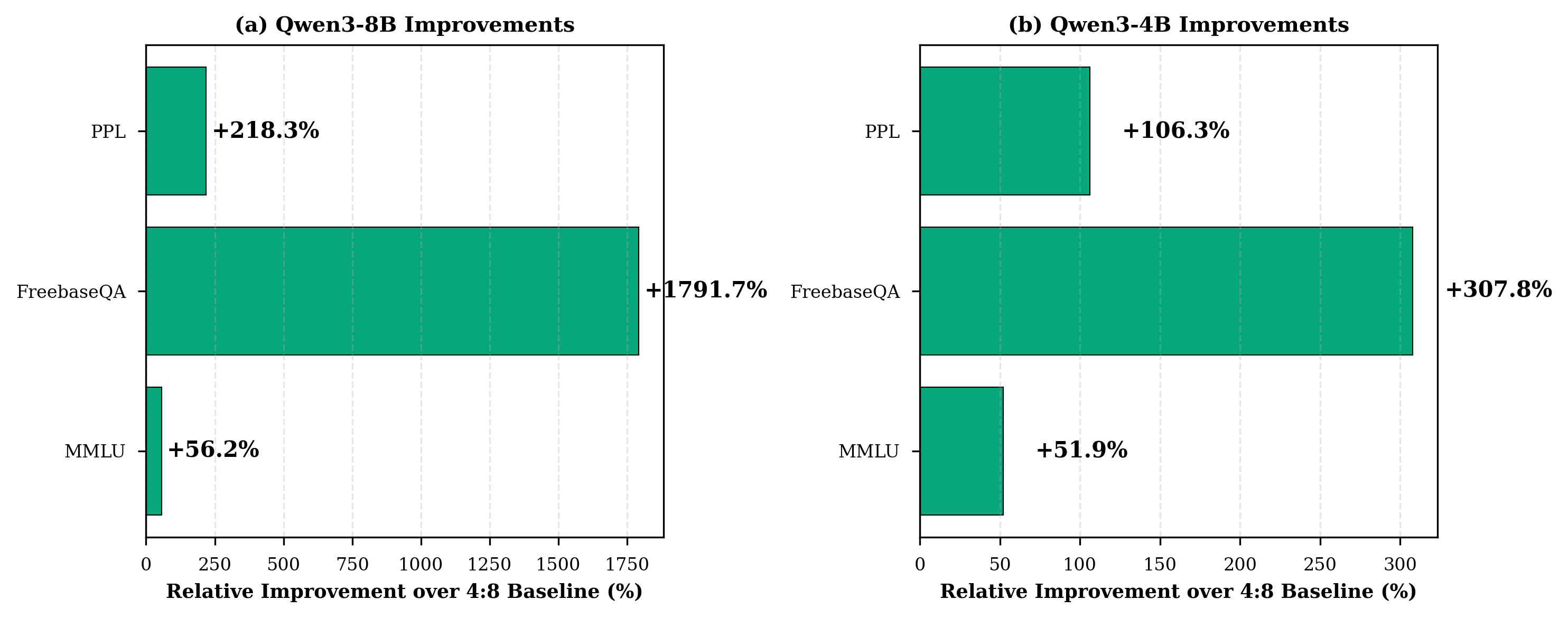}
		\caption{Relative improvements of our agent-guided method over the best structured baseline (4:8) for both model sizes. Positive percentages indicate superior performance; for perplexity, we show the reduction in degradation (higher is better). The consistent improvements across models and metrics demonstrate the robustness of our approach.}
		\label{fig:relative_improvements}
	\end{figure}
		\begin{table}[t]
		\caption{Iteration statistics for agent-guided pruning. The low rollback rates (9.5-10\%) demonstrate effective self-correction through the reflection mechanism, with both models successfully reaching the 50\% target sparsity.}\label{tab:iteration_stats}
		\centering
		\small
		\begin{tabular}{lcc}
			\toprule
			\textbf{Statistic} & \textbf{Qwen3-8B} & \textbf{Qwen3-4B} \\
			\midrule
			Total Iterations & 21 & 40 \\
			Successful Iterations & 19 & 36 \\
			Rollbacks & 2 & 4 \\
			Rollback Rate & 9.5\% & 10.0\% \\
			\midrule
			Target Sparsity & 50.0\% & 50.0\% \\
			Final Sparsity & 50.73\% & 50.46\% \\
			Target Achievement & 101.5\% & 100.9\% \\
			\midrule
			Baseline PPL & 22.06 & 27.90 \\
			Final PPL & 29.49 & 53.09 \\
			PPL Degradation & +33.7\% & +90.3\% \\
			\bottomrule
		\end{tabular}
	\end{table}

	\subsection{Agent Behavior Analysis}
	
	Table~\ref{tab:iteration_stats} summarizes the iteration statistics for both models. The Qwen3-8B model reached 50.73\% sparsity in 21 iterations with only 2 rollbacks (9.5\% rollback rate), while Qwen3-4B reached 50.46\% sparsity in 40 iterations with 4 rollbacks (10\% rollback rate). These low rollback rates indicate that the agent makes predominantly correct decisions, with the self-reflection mechanism enabling effective learning across iterations.

	Figures~\ref{fig:cumulative_8b} and~\ref{fig:cumulative_4b} visualize the iterative pruning process for both models. The top panels show the evolution of sparsity (left) and perplexity (right) across iterations. The orange lines represent the state before each pruning decision, while green lines show the state after pruning. The close tracking of these lines demonstrates that the agent makes small, incremental adjustments rather than large, risky changes. The bottom panels show per-iteration statistics: sparsity gain achieved (left) and perplexity change incurred (right). 
	
	For Qwen3-8B, we observe that the agent learns to maintain steady progress toward the target, with most iterations achieving 1-3\% sparsity gains while keeping perplexity changes below 2\%. The two rollback events (visible as gaps in the bottom panel) occur at iterations 17 and 20, when the agent becomes too aggressive and triggers the 15\% PPL threshold. After each rollback, the agent adjusts its strategy to be more conservative.
	
	For Qwen3-4B, the pruning trajectory shows similar patterns but with more iterations required to reach the target. The agent achieves particularly large sparsity gains (3-9\%) in early iterations (2-3) when the model is relatively robust, then becomes more conservative as perplexity begins to rise. The four rollback events are distributed across iterations 10, 15, 25, and 32, each followed by visible strategy adjustments in subsequent iterations.
	
	These visualizations reveal several key behaviors: (1) the agent learns to "front-load" pruning early when the model is most robust, (2) it becomes progressively more conservative as cumulative perplexity degradation increases, (3) rollbacks are rare and trigger meaningful strategy changes, and (4) the agent successfully navigates the trade-off between sparsity gain and model quality preservation throughout the pruning process.

	The agent's reasoning reveals several learned patterns: (1) prioritizing layers with highly negative sensitivity z-scores (typically below -1.0), (2) avoiding layers with positive gradient z-scores indicating high loss impact, (3) becoming more conservative as perplexity degrades, and (4) accelerating pruning when perplexity remains stable. The self-reflection feedback enables the agent to recognize when it has been too aggressive (high perplexity increase) or too conservative (minimal sparsity gain), adjusting its strategy accordingly.
	
	\subsection{Visualization}
	
	Figures~\ref{fig:comparison_8b} and~\ref{fig:comparison_4b} visualize the performance comparison across all three metrics for both model sizes. The dramatic differences in FreebaseQA—where structured methods experience near-total collapse—clearly illustrate the severity of the factual knowledge degradation problem and our method's effectiveness in addressing it.
	
	Figure~\ref{fig:mmlu_categories} shows the MMLU category breakdown for Qwen3-8B, demonstrating that our approach preserves capabilities across diverse knowledge domains rather than exhibiting selective preservation in specific areas. Figure~\ref{fig:relative_improvements} presents the relative improvements over the 4:8 baseline, highlighting our method's consistent advantages across metrics.

	\section{Conclusion}
	
	We introduced agent-guided pruning, a novel framework where foundation models adaptively compress other foundation models through iterative reasoning and self-reflection. By constructing layer-wise sensitivity profiles that combine Wanda-inspired weight-activation metrics with gradient importance scores, normalized as z-scores for model-agnostic comparison, we enable an LLM agent to intelligently select which layers to prune at each iteration while learning from previous outcomes through a self-reflection mechanism. We evaluated our approach on Qwen3 models (4B and 8B parameters) at approximately 45\% sparsity, demonstrating substantial improvements over structured pruning baselines: 56\% relative improvement in MMLU accuracy, 19× better factual knowledge retention on FreebaseQA, and 69\% lower perplexity degradation compared to 4:8 structured pruning. With only 9.5-10\% rollback rates across 21-40 iterations, our framework exhibits effective self-correction without requiring retraining or manual heuristic design. These results directly address the critical factual knowledge collapse problem identified by recent benchmarks, establishing that foundation models can effectively guide neural network compression in ways that preserve capabilities missed by traditional metrics.

\end{document}